\documentclass{article}

\usepackage{palatino}

\usepackage{microtype}
\usepackage{graphicx}
\usepackage{booktabs}
\usepackage{textcomp}

\usepackage{amsmath,amssymb} 
\usepackage{color}

\usepackage[hidelinks]{hyperref}
\usepackage{multirow}
\usepackage{bbm}

\usepackage{tikz}
\usepackage{comment}

\usepackage{algorithm}
\usepackage{algorithmic}

\newcommand{\magenta}[1]{{\color{magenta} #1}}

\usepackage{axessibility}  

\usepackage[font=small,labelfont=bf]{caption}
\usepackage{subcaption}

\begin{document}

\title{Localizing Visual Sounds the Easy Way} 

\author{Shentong Mo \and Pedro Morgado}
\date{Carnegie Mellon University\\Pittsburgh PA 15213, USA}

\maketitle

\begin{abstract}
\noindent Unsupervised audio-visual source localization aims at localizing visible sound sources in a video without relying on ground-truth localization for training. Previous works often seek high audio-visual similarities for likely positive (sounding) regions and low similarities for likely negative regions. However, accurately distinguishing between sounding and non-sounding regions is challenging without manual annotations. In this work, we propose a simple yet effective approach for Easy Visual Sound Localization, namely EZ-VSL, without relying on the construction of positive and/or negative regions during training. Instead, we align audio and visual spaces by seeking audio-visual representations that are aligned in, at least, one location of the associated image, while not matching other images, at any location. We also introduce a novel object guided localization scheme at inference time for improved precision. Our simple and effective framework achieves state-of-the-art performance on two popular benchmarks, Flickr SoundNet and VGG-Sound Source. In particular, we improve the CIoU of the Flickr SoundNet test set from 76.80\% to 83.94\%, and on the VGG-Sound Source dataset from 34.60\% to 38.85\%. The code is available at \href{https://github.com/stoneMo/EZ-VSL}{\magenta{https://github.com/stoneMo/EZ-VSL}}.
\end{abstract}

\section{Introduction}
\label{sec:intro}

When we hear a baby crying, we can localize the sound by finding the baby in the room. This ability of visual sound source localization is possible due to the tight association between visual and auditory signals in the natural world. 
In this work, we aim to leverage this natural and freely available audio-visual association to localize sound sources present in a video in an unsupervised manner, i.e.~without relying on manual annotations for sounding source locations.

Unsupervised visual localization of sound sources has attracted much attention in recent years~\cite{Senocak2018learning,Afouras2020selfsupervised,chen2021localizing}.
To tackle this problem, recent approaches~\cite{hu2019deep,Afouras2020selfsupervised,qian2020multiple,chen2021localizing,arda2022learning} rely on direct audio-visual similarity in a learned latent space for localization.
These audio-visual similarities are used to construct likely sounding and non-sounding regions in the image, and the models are learned by requiring the audio representation to match visual representations pooled from likely sounding regions while being dissimilar from those of different images~\cite{hu2019deep,Afouras2020selfsupervised,qian2020multiple}, and/or from non-sounding regions~\cite{chen2021localizing,arda2022learning}. While these approaches have been shown to yield state-of-the-art performance in unsupervised visual sound localization, we identify two major limitations.

First, the training objective presents a paradox. On one hand, accurate regions of sounding objects are required in order to encourage audio representations to match the visual representations of the regions where the source is located. On the other hand, since localization maps are obtained through audio-visual similarities, accurate representations are required in order to identify the regions containing the sounding objects. This paradox results in a complex training objective that is likely to contain many sub-optimal local minima, as the model is required to bootstrap from its own localization ability.

Second, by solely relying on audio-visual similarity for localization, prior work ignores the visual prior of likely audio sources. For example, even without access to the audio signal, we know that most regions of an image, depicting for example the floor, the sky, a table, or a wall, are unlikely to depict sources of sound.

\begin{figure}[t]
\centering
\begin{subfigure}{0.45\linewidth}
    \centering
    \includegraphics[width=\linewidth]{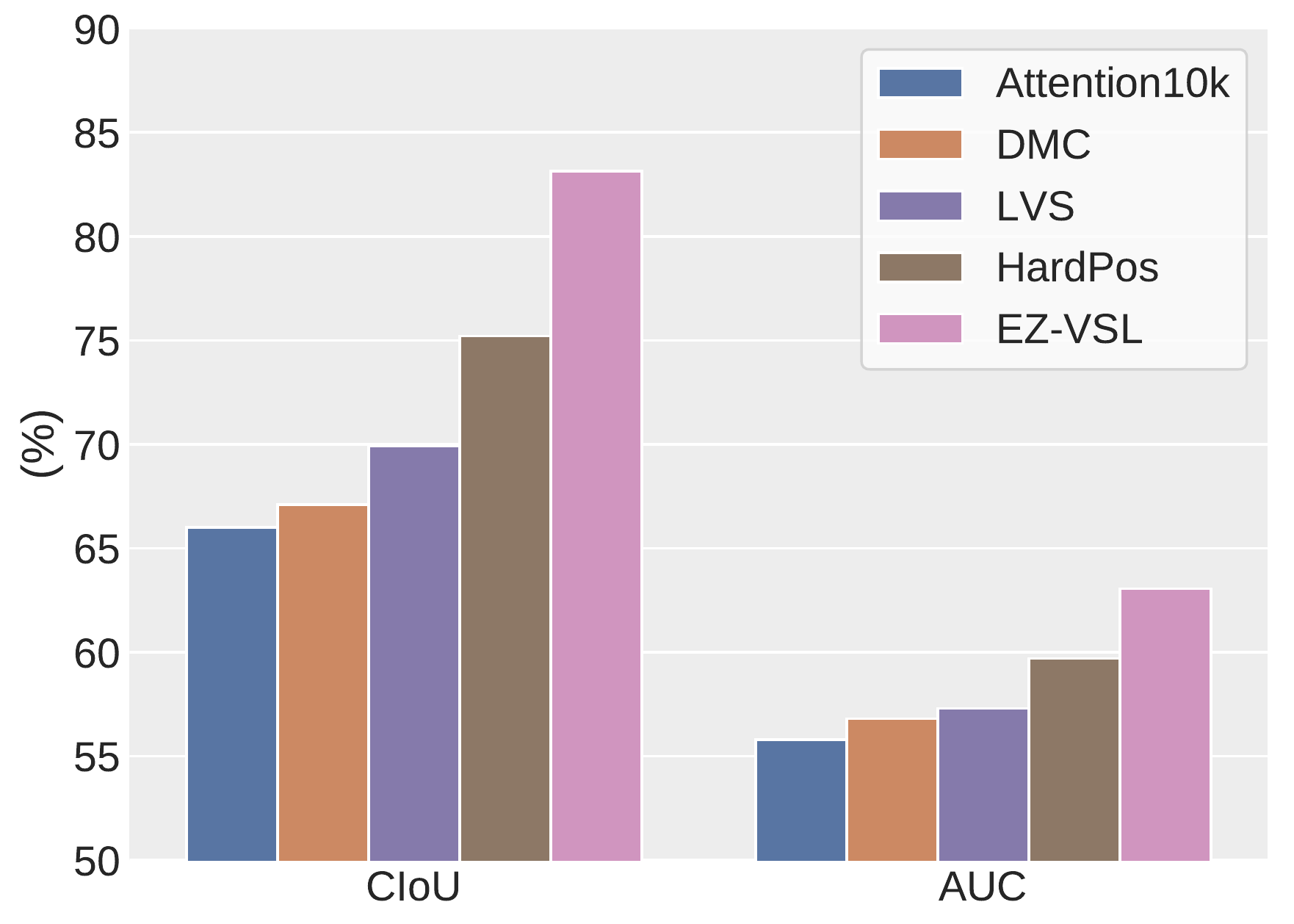}
    \caption{Flick SoundNet}
\end{subfigure}%
\begin{subfigure}{0.45\linewidth}
    \centering
    \includegraphics[width=\linewidth]{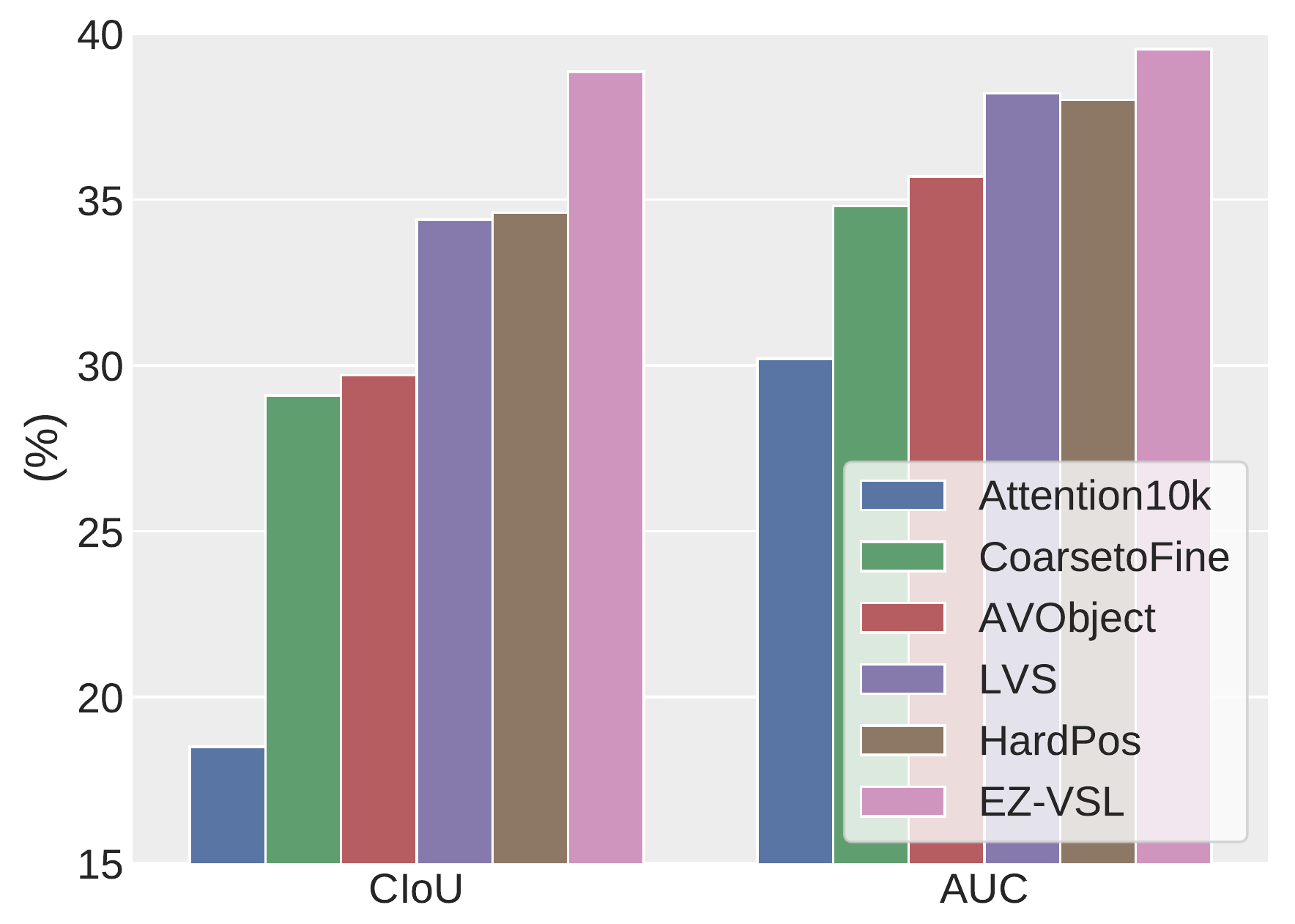}
    \caption{VGG Sound Sources}
\end{subfigure}
\caption{Comparison of EZ-VSL with state-of-the-art methods on Flickr SoundNet~\cite{hu2019deep} (a) and VGG-SS~\cite{chen2021localizing}. All methods in (a) are trained on Flickr 144k, and those in (b) on VGG Sound 144k.}
\label{fig: title_img}
\end{figure}

To address these challenges, we propose a simple yet effective approach for easy visual sound localization, namely EZ-VSL.
Instead of relying on explicit maps for sounding and non-sounding regions, we treat audio-visual correspondence learning as a multiple instance learning problem. In other words, we propose a training loss that encourages the audio signal to be associated with, at least, one location in the corresponding image, while not being associated with any location from other images. 
Then, we introduce a novel object-guided localization scheme at inference time that combines the audio-visual similarity map with an object localization map from a lightweight pre-trained visual model, which biases sound source localization predictions towards the objects in the scene. 


We evaluate our EZ-VSL on two popular benchmarks, Flickr SoundNet~\cite{hu2019deep} and VGG-Sound Source~\cite{chen2021localizing}.
Extensive experiments show the superiority of our approach for unsupervised sound source visual localization.
We also conduct comprehensive ablation studies to demonstrate the effectiveness of each component.
Surprisingly, we found that the object prior alone, which does not even leverage the audio for localization, already surpasses all prior work on both Flickr and VGG-Sound benchmarks.
We also demonstrate the superiority of the proposed multiple instance learning objective for audio-visual matching compared to prior approaches that rely on careful constructions of positive (sounding) and negative (non-sounding) regions for training.
Finally, we show that the visual object prior and audio-visual similarity maps can be further combined into more accurate predictions, surpassing the current state-of-the-art method by large margins on both Flickr SoundNet and VGG Sound Sources. These results are highlighted in Fig.~\ref{fig: title_img}.


Overall, the main contributions of this work can be summarized as follows:
\begin{itemize}
    \item We present a simple yet effective multiple instance learning framework for unsupervised sound source visual localization, which we call EZ-VSL.
    \item We propose a novel object-guided localization scheme that favors object regions, which are more likely to contain sound sources.
    \item Our EZ-VSL successfully achieves state-of-the-art performance on two popular benchmarks, Flickr SoundNet and VGG-Sound Source.
\end{itemize}

\section{Related Work}
\label{sec:related_work}

\subsection{Audio-Visual Joint Learning}

In recent years, many works~\cite{aytar2016soundnet,owens2016ambient,Arandjelovic2017look,wang2017untrimmednets,korbar2018cooperative,Morgado2018self,tian2020unified,Morgado2020learning,Morgado2021robust,Morgado2021audio} have been proposed on audio-visual joint learning to learn bimodal representations from each other. 
SoundNet~\cite{aytar2016soundnet} applies a visual teacher network to extract audio representations from untrimmed videos.
The audio-visual correspondence task~\cite{Arandjelovic2017look} is introduced to learn both visual and audio representations in an unsupervised way.
Audio-visual synchronization objectives are also explored for several tasks, such as speech recognition~\cite{Chung2017lip,Afouras2018DeepLR}, audio-visual navigation~\cite{Chen2020SoundSpacesAN}, visual sound source separation, and localization~\cite{Gan2020music,Gao2018learning,Gao2019co,Senocak2018learning,zhao2018the,zhao2019the}.

Besides these works, several methods adopt a weakly-supervised scheme to solve audio-visual problems. For example, UntrimmedNet~\cite{wang2017untrimmednets} uses a classification module and a selection module for Multiple Instance Learning (MIL) to perform audio-visual action localization. \cite{tian2020unified} also proposes in a hybrid attention network for audio-visual video parsing. In this work, however, we focus on the sound source localization problem by learning audio-visual representations jointly from unlabelled videos.

\subsection{Audio-Visual Source Localization}
Audio-Visual Source Localization aims at localizing sound sources by learning the co-occurrence of audio and visual features in a video.
Early works~\cite{hershey1999audio,fisher2000learning,kidron2005pixels} use shallow probabilistic models or canonical correlation analysis to solve this problem.
With the introduction of deep neural networks, 
some approaches~\cite{hu2019deep,Owens2018audio} were proposed to learn the audio-visual correspondence via a dual-stream network and a contrastive loss.
For instance, DMC~\cite{hu2019deep} adopts synchronous sets of clustering with respect to each modality for capturing audio-visual correspondences.
Multisensory features~\cite{Owens2018audio} are used to jointly learn visual and audio representations of a video through the temporal alignment.
Other methods~\cite{zhao2018the,zhao2019the,Rouditchenko2019SelfsupervisedAC,gao20192.5D,gan2019self} leverage the audio-visual source separation as the target to achieve visual sound localization. Most of these methods learn from global audio-visual correspondences. Although they show qualitatively that the model is capable of localization, their localization ability is not competitive to models that learn from localized correspondences.

Beyond the work discussed above, several relevant works have targeted the visual source localization problem directly. 
Attention10k~\cite{Senocak2018learning} developed an attention mechanism and a two-stream architecture with each modality to localize sound sources in an image.
Qian \textit{et al.}~\cite{qian2020multiple} proposed a two-stage framework to learn audio and visual representations with the cross-modal feature alignment in a coarse-to-fine way. 
Afouras \textit{et al.}~\cite{Afouras2020selfsupervised} introduced an attention-based model with the optical flow to localize and group sound sources in a video.
More recently, LVS~\cite{chen2021localizing} added a hard sample mining mechanism to contrastive loss with a differentiable threshold on the audio-visual correspondence map.
Finally, HardPos~\cite{arda2022learning} leveraged hard positives in contrastive learning for learning semantically matched audio-visual information from negative pairs.
Different from these baselines, we show that it is possible (and even preferable) to learn from a simplified multiple-instance contrastive learning objective.
Furthermore, we propose a novel object guided localization scheme to boost the visual localization performance of sound sources.

\begin{figure*}[!tb]
    \centering
    \includegraphics[width=0.8\linewidth]{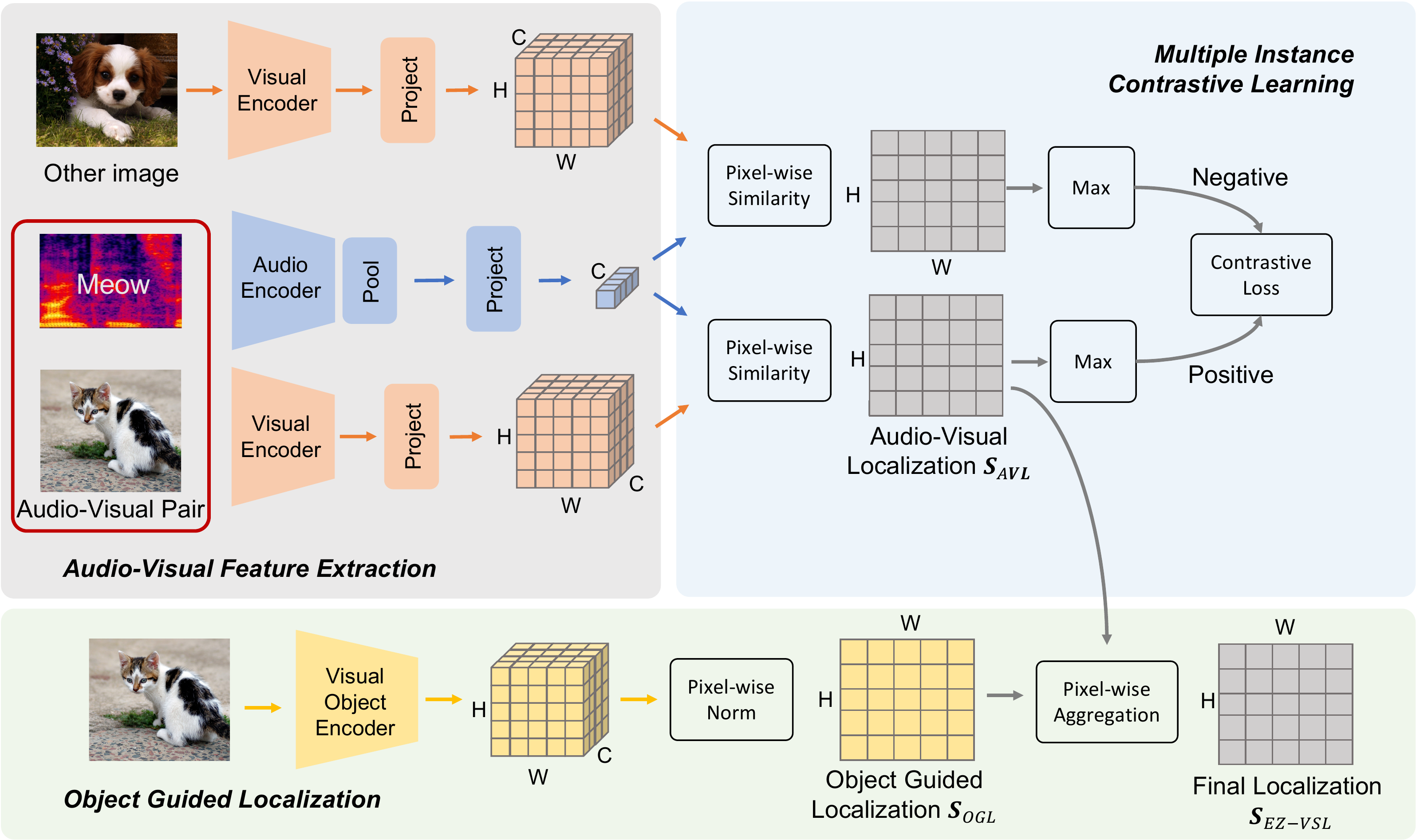}
    \caption{Illustration of the proposed easy visual sounds localization (EZ-VSL). The audio-visual feature extractor computes global audio and localization visual features. Audio-visual alignment is learned using a multiple instance contrastive learning objective. 
    At inference time, we use another visual encoder pre-trained on object recognition to compute object localization maps, which are combined with audio-visual localization maps for the final prediction.}
	\label{fig: main_img}
\end{figure*}

\section{Method}
\label{sec:method}
Given a video containing sound sources, our goal is to localize the sounding objects within it without using manual annotations of their locations for training.
We propose a simple yet effective way for unsupervised sound source visual localization, which we denote EZ-VSL.

\subsection{Overview}
Let $\mathcal{D}=\{(v_i, a_i): i=1, \ldots, N\}$ be a dataset of paired audio $a_i$ and visual data $v_i$, where the sources of the sound audible in $a_i$ are assumed to be depicted in $v_i$.
Following previous work~\cite{hu2019deep,chen2021localizing}, we first encode the audio and visual signals using a two stream neural net encoder, denoted as $f_a(\cdot)$ and $f_v(\cdot)$ for the audio and images, respectively. The audio encoder extracts global audio representations $\mathbf{a}_i=f_a(a_i)$ and the visual encoder computes localized representations $\mathbf{v}_i^{xy}=f_v(v_i^{xy})$ for each $(x,y)$ location.
As shown in Fig.~\ref{fig: main_img}, audio and visual features are then mapped into a shared latent space, where the similarity between audio-visual representations can be computed for all locations. The audio-visual models are then trained to minimize a \textbf{cross-modal multiple-instance contrastive learning} loss, that encourages audio representation to be aligned with the associated visual representations at least at one location. By optimizing this loss, audio and visual signals are matched in the shared latent space, which can then be used for localization.
At inference time, we combine the learned audio-visual  similarities with \textbf{object guided localization}. We accomplish this using a visual model pre-trained for object recognition. It should be noted that models pre-trained on ImageNet are already used to initialize the visual encoder for VSL~\cite{Senocak2018learning,hu2019deep,Afouras2020selfsupervised,qian2020multiple,chen2021localizing,arda2022learning}. We use the same model to extract regions of the image that are likely to contain objects (regardless of whether they are producing the sound or not). The object maps are then integrated with audio-visual similarities to enhance localization accuracy.

We now elaborate on the two main components of our work: the multiple instance contrastive learning objective, and the object guided localization.

\subsection{Audio-visual matching by multiple-instance contrastive learning}
Aligning audio and localized visual representations poses two main challenges. First, the output of the audio and visual encoders are not necessarily compatible. Second, most locations in the image do not depict the sound source, and so the representations at these locations should not be aligned with the audio.

The first challenge  can be easily addressed by projecting both audio and visual representations into a shared feature space
\begin{equation}
\label{eq: first}
    \hat{\mathbf{v}}_i^{xy} = \mathbf{U}_v \mathbf{v}_i + \mathbf{b}_v \quad \forall x,y,i \qquad \mbox{and} \qquad
    \hat{\mathbf{a}}_i = \mathbf{U}_a \mathbf{a}_i + \mathbf{b}_a \quad \forall i
\end{equation}
where $\mathbf{U}_v$ and $\mathbf{U}_a$ are projection matrices, and $\mathbf{b}_v$ and $\mathbf{b}_a$ bias terms.
While prototyping the model, we experimented with non-linear projections, but we did not observe significant improvements.

The second challenge requires to selectively match the audio representations to the associated visual regions depicting the sound sources. Prior work~\cite{Senocak2018learning,hu2019deep,qian2020multiple,Afouras2020selfsupervised,chen2021localizing,arda2022learning} explicitly computes an attention map for the likely sounding regions by bootstrapping from current audio-visual similarities.
The audio representations are then required to match these sounding regions~\cite{Senocak2018learning,hu2019deep,qian2020multiple}, and in some cases to not match non-sounding regions from the same image~\cite{chen2021localizing,arda2022learning}. As discussed above, this causes a paradox where accurate localizations are required to learn accurate audio-visual representations required for localization in the first place. 

To simplify this framework, we propose to optimize a multiple instance contrastive learning loss. Each bag of visual features $V$ spans all locations within an image
\begin{equation}
    V_i=\{\hat{\mathbf{v}}_i^{xy}: \forall x, y\} \quad \forall i\in\mathcal{D}
\end{equation}
Audio representations $\mathbf{a}_i$ are then required to be similar to at least one instance in the corresponding positive bag $V_i$, while being dissimilar from all locations in all negative bags $V_j\ \forall j\neq i$. Following the spirit of contrastive learning approaches, negative bags are obtained from other images in the same mini-batch.
More specifically, we seek to maximize the alignment between the audio and the most similar positive visual instance, through the following loss function
\begin{equation}
    \label{eq:micl}
    \mathcal{L}_{a \rightarrow v} = - \log \frac{
    \exp \left( \frac{1}{\tau} \max_{\hat{\mathbf{v}}\in V_i} \mathtt{sim}(\hat{\mathbf{a}}_i, \hat{\mathbf{v}}) \right)
    }{
    \sum_k \exp \left(  \frac{1}{\tau} \max_{\hat{\mathbf{v}}\in V_k} \mathtt{sim}(\hat{\mathbf{a}}_i, \hat{\mathbf{v}})\right)}  
\end{equation}
where $\mathtt{sim}(\hat{\mathbf{v}},\hat{\mathbf{a}}) = \hat{\mathbf{v}}^T\hat{\mathbf{a}} / (\|\hat{\mathbf{v}}\| \|\hat{\mathbf{a}}\|) $ is the cosine similarity, and $\tau$ a temperature hyper-parameter.
To train our models, we use a symmetric version of (\ref{eq:micl}) by defining
\begin{equation}
    \mathcal{L}_{v \rightarrow a} = - \log \frac{
    \exp \left( \frac{1}{\tau} \max_{\hat{\mathbf{v}}\in V_i} \mathtt{sim}(\hat{\mathbf{v}}, \hat{\mathbf{a}}_i) \right)
    }{
    \sum_k \exp \left(  \frac{1}{\tau} \max_{\hat{\mathbf{v}}\in V_i} \mathtt{sim}(\hat{\mathbf{v}}, \hat{\mathbf{a}}_k) \right)},
\end{equation}
and optimizing the symmetric loss
\begin{equation}
    \label{eq:micl_sym}
    \mathcal{L} = \mathcal{L}_{a \rightarrow v} + \mathcal{L}_{v \rightarrow a}.
\end{equation}

During inference, direct cosine similarity is used to compute the audio-visual localization map
\begin{equation}
    \label{eq:s_avl}
    \mathbf{S}_{xy}^{AVL} = \mathtt{sim}(\hat{\mathbf{v}}_{xy}, \hat{\mathbf{a}}) \quad \forall x \in [1,W], y \in [1,H].
\end{equation}

\subsection{Object-Guided Localization}
At inference time, we propose a novel object-guided scheme for enhanced localization. 
The input image is fed to a convolutional model $f_{obj}$ pre-trained on ImageNet~\cite{imagenet_cvpr09} without global pooling or the classification head, yielding a feature map $\mathbf{v}^\prime = f_{obj}(v)\in\mathbb{R}^{C \times H \times W}$. This model has the same architecture than the visual encoder used for audio-visual localization and is initialized with the same ImageNet pre-trained weights, but unlike the former, this model is never trained for audio-visual similarity. Hence, the feature map $\mathbf{v}^\prime$ contains zero information about the accompanying audio. Instead, it can be used to define a localization prior that favors the objects in the scene, regardless of whether these objects are the sources of the sound or not. We then experimented with two possible solutions to extract object-centric localization maps without any additional training. The first uses the ImageNet pretrained classifier to obtain a 1000-way object class posterior at each $(x, y)$ location $P(o|\mathbf{v}^\prime_{xy})$, and defines the object localization prior as 
\begin{equation}
    \label{eq:s_cls}
    \mathbf{S}_{xy}^{CLS} = \max_o P(o|\mathbf{v}^\prime_{xy}).
\end{equation}
The second approach, perhaps less intuitive but more effective, relies on the fact that $f_{obj}$ was trained on an object-centric dataset, and thus produces stronger activations when evaluated on images of objects. With this intuition in mind, we alternatively define the object localization prior as 
\begin{equation}
    \label{eq:s_l1}
    \mathbf{S}_{xy}^{L1} = \|\mathbf{v}^\prime_{xy}\|_1.
\end{equation}

The audio-visual localization and object-centric maps are then linearly aggregated into a final localization map $\mathbf{S}_{xy}^{EZVSL}$ of the form
\begin{equation}
    \label{eq:s_ezvsl}
    \mathbf{S}_{xy}^{EZVSL} = \alpha \mathbf{S}_{xy}^{AVL}+ (1-\alpha)\mathbf{S}_{xy}^{OBJ} \quad \forall x,y,
\end{equation}
where $\mathbf{S}_{xy}^{AVL}$ is the audio-visual similarity of map of (\ref{eq:s_avl}), $\mathbf{S}_{xy}^{OBJ}$ is the object localization map (i.e., $\mathbf{S}_{xy}^{CLS}$ in (\ref{eq:s_cls}) or $\mathbf{S}_{xy}^{L1}$ in (\ref{eq:s_l1})), and $\alpha$ is balancing term that weights the contribution of the object prior and the audio-visual similarity terms.
In practice, since the two maps $\mathbf{S}_{xy}^{AVL}$ and $\mathbf{S}_{xy}^{OBJ}$ can have widely different ranges of scores, we normalize them into a $[0, 1]$ range before aggregation, i.e., $\mathbf{S}_{xy}=\frac{\mathbf{S}_{xy}-\min_{xy}\mathbf{S}_{xy}}{\max_{xy}\mathbf{S}_{xy}-\min_{xy}\mathbf{S}_{xy}}$.

\section{Experiments}
\label{sec:experiments}
We evaluated EZ-VSL on unsupervised visual sound source localization. Following accepted practices~\cite{Senocak2018learning,qian2020multiple,chen2021localizing}, we used the Flickr SoundNet dataset~\cite{aytar2016soundnet} and the recently proposed VGG-Sound dataset~\cite{chen2020vggsound}, and report the same evaluation metrics as in~\cite{Senocak2018learning,qian2020multiple,chen2021localizing}. Namely, we measure the average precision at a Consensus Intersection over Union threshold of $0.5$, a metric often simply denoted as CIoU. We also measure the Area Under Curve (AUC).

\subsection{Experimental setup}

\paragraph{\textbf{Datasets}}
Flickr SoundNet includes 2 million unconstrained videos from Flickr. 
From each video clip, a single image frame is extracted together with 20s of audio centered around it, to form the corresponding audio-visual pairs used for unsupervised learning. 
We also conduct experiments on \textbf{VGG-Sound} composed of 200k video clips from 309 sound categories. 
Similar to the Flickr dataset, the video is represented by a single frame as well as its audio. 
To enable direct comparisons with existing work~\cite{chen2021localizing,Senocak2018learning,qian2020multiple,Afouras2020selfsupervised}, we trained our models using subsets of either 10k or 144k image-audio pairs.

Localization performance is measured on two datasets, the Flickr SoundNet test set~\cite{Senocak2018learning} and the more challenging VGG-Sound Sources test set~\cite{chen2021localizing}. The former includes only 250 image-audio pairs for which the location of the sound source has been manually annotated. The latter contains annotations for 5000 instances spanning 220 sounding objects categories.

\paragraph{Audio and visual pre-processing}
The input to the visual encoder $f_v(\cdot)$ are images of resolution $224 \times 224$. During training, images are first resized to $246$ along the shortest edge, and random cropping together with random horizontal flipping is applied for data augmentation. At test time, images directly resized into a $224 \times 224$ resolution without cropping.

The audio encoder $f_a(\cdot)$ takes the log spectrograms extracted from $3s$ of audio as the input, resulting in an input tensor of size $257 \times 300$ ($257$ frequency bands over $300$ timesteps). No data augmentations are applied during train or test time.

\paragraph{\textbf{Audio and visual models}}
Both the visual and audio encoders are implemented using the lightweight ResNet18~\cite{he2016resnet} as the backbone. 
Following prior work~\cite{hu2019deep,qian2020multiple,chen2021localizing}, we initialized the visual model using weights pre-train on ImageNet~\cite{imagenet_cvpr09}.
Unless otherwise specified, the audio and visual representations are projected into a shared space of dimension $512$. 

The model is trained with a batch size of 128 on 2 GPUs. For efficiency, we only use negatives from the local batch, i.e.~we did not gather negatives from all GPUs. This results in a negative set of 63 samples for the contrastive learning objective of (\ref{eq:micl}). The model is trained using the Adam optimizer~\cite{kingma2014adam} with a learning rate of $1e-4$, and default hyper-parameters $\beta_1=0.9, \beta_2=0.999$. On large datasets (144k or the full VGG-Sound database), the model is trained for 20 epochs. On smaller (10k) datasets, the model is trained for 100 epochs. 

\begin{table}[!tb]
	\renewcommand\tabcolsep{6.0pt}
	\centering
	\caption{Comparison results on Flickr SoundNet testset where models are trained on Flickr 10k and 144k data. }
	\label{tab: exp_sota_flickr}
	\scalebox{0.85}{
		\begin{tabular}{llcc}
			\toprule
			Training set & Method & CIoU(\%) & AUC(\%) \\ 	
			\midrule
			\multirow{5}{*}{Flickr 10k} & Attention10k~\cite{Senocak2018learning} & 43.60& 44.90 \\
			 & CoarsetoFine~\cite{qian2020multiple} & 52.20 & 49.60 \\
			 & AVObject~\cite{Afouras2020selfsupervised} & 54.60 & 50.40 \\
			 & LVS~\cite{chen2021localizing} & 58.20 & 52.50 \\
			 & EZ-VSL (ours) & \textbf{81.93} & \textbf{62.58} \\ \hline
			 \multirow{4}{*}{Flickr 144k} & Attention10k~\cite{Senocak2018learning} & 66.00 & 55.80 \\
			 &  DMC~\cite{hu2019deep} & 67.10 & 56.80 \\
			 & LVS~\cite{chen2021localizing} & 69.90 & 57.30 \\
			 & HardPos~\cite{arda2022learning} & 75.20 & 59.70 \\
			 & EZ-VSL (ours) & \textbf{83.13} & \textbf{63.06} \\
			\bottomrule
			\end{tabular}}
\end{table}

\begin{table}[!tb]
	\renewcommand\tabcolsep{6.0pt}
	\centering
	\caption{Comparison results on Flickr SoundNet and VGG-SS testset where models are trained on VGG-Sound 144k data. }
	\label{tab: exp_sota_vggss}
	\scalebox{0.85}{
		\begin{tabular}{llccccc}
			\toprule
			\multirow{2}{*}{Training set} &
			\multirow{2}{*}{Method} & \multicolumn{2}{c}{Flickr-SoundNet} & \multicolumn{2}{c}{VGG-SS} \\
			&& CIoU(\%) & AUC(\%) & CIoU(\%) & AUC(\%) \\ 	
			\midrule
			\multirow{5}{*}{VGG-Sound 144k} &
			Attention10k~\cite{Senocak2018learning} & 66.00 & 55.80 & 18.50 & 30.20 \\
			& CoarsetoFine~\cite{qian2020multiple} & - & - &  29.10 & 34.80 \\
			& AVObject~\cite{Afouras2020selfsupervised} & - & - & 29.70 & 35.70\\
			& LVS~\cite{chen2021localizing} & 73.50 & 59.00 & 34.40 & 38.20 \\
			& HardPos~\cite{arda2022learning} & 76.80 & 59.20 & 34.60 & 38.00 \\
			& EZ-VSL (ours) & \textbf{83.94} & \textbf{63.60} & \textbf{38.85} & \textbf{39.54} \\
			\bottomrule
			\end{tabular}}
\end{table}

\subsection{Comparison to prior work}
In this work, we propose a simple yet highly effective training framework for visual sound source localization. 
To demonstrate the effectiveness of our approach, EZ-VSL,  we start by drawing direct comparisons to previous works~\cite{Senocak2018learning,qian2020multiple,Afouras2020selfsupervised,chen2021localizing} on two popular benchmarks: Flickr SoundNet~\cite{hu2019deep} and VGG-SS~\cite{chen2021localizing}. Results are reported in Tables~\ref{tab: exp_sota_flickr} and \ref{tab: exp_sota_vggss} for models trained on Flickr SoundNet and VGG-SS, respectively.

As can be seen, EZ-VSL outperforms prior work by large margins, establishing new state-of-the-art results in all settings. 
On the Flickr test set, we observe performance gains of 23.73\% CIoU and 10.08\% AUC when models are trained on Flickr 10k, by 7.93\% CIoU and 3.36\% AUC when trained on Flickr 144k, and by 7.14\% CIoU and 4.4\% AUC when trained on VGG-Sound 144k.
Significant gains can also be observed on the more challenging VGG-Sound Sources test set, with EZ-VSL outperforming prior work by 4.25\% CIoU and 1.34\% AUC.

We highlight that these gains are obtained with a significantly simplified training objective. For example, Attention10K~\cite{Senocak2018learning} relies on the construction of positive (sounding) regions for its visual attention mechanism, and both LVS~\cite{chen2020vggsound} and HardPos~\cite{arda2022learning} require not only the construction of likely positive (sounding) regions but also negative (non-sounding) regions. This highlights the importance of a well-designed training framework that avoids imposing complex region-specific constraints. 
Also, note that our method combines both the novel multiple instance contrastive learning loss used for training and the novel object-centric localization procedure used during inference. The effect of these individual components will be studied below.

\subsection{Open Set Audio-Visual Localization}
To assess generalization, we evaluated the ability of EZ-VSL to generalize beyond the categories of sound sources heard during self-supervised training. 
Following previous work~\cite{chen2021localizing}, we randomly sampled 110 categories from VGG-Sound for training. We then evaluate our model on test samples from these heard categories, as well as on samples from another 110 unheard categories.
The results are shown in Table~\ref{tab: exp_openset}. As can be seen, our approach outperforms LVS~\cite{chen2021localizing} by a significant margin on both heard and unheard categories. In fact, unlike LVS, the performance of our EZ-VSL model did not suffer by the presence of unheard sound categories, achieving even slightly better performance on unheard classes than on heard classes. This provides evidence for the stronger generalization ability of EZ-VSL in an open set setting.

\begin{table}[t]
	\renewcommand\tabcolsep{6.0pt}
	\centering
	\caption{Comparison results on VGG-SS for open set audio-visual localization trained on 70k data with heard 110 classes.}
	\label{tab: exp_openset}
	\scalebox{0.9}{
		\begin{tabular}{clcc}
			\toprule
			Test class & Method & CIoU(\%) & AUC(\%)  \\ 	
			\midrule
             \multirow{2}{*}{Heard 110} & LVS~\cite{chen2021localizing} & 28.90 & 36.20 \\
             & EZ-VSL & \textbf{37.25} & \textbf{38.97} \\\midrule
             \multirow{2}{*}{Unheard 110} & LVS~\cite{chen2021localizing} & 26.30 & 34.70 \\
             & EZ-VSL & \textbf{39.57} & \textbf{39.60} \\
             \bottomrule
			\end{tabular}}
\end{table}

\subsection{Cross Dataset Generalization}
To further evaluate generalization, we tested models across datasets. Specifically, we tested the model trained on VGG-Sound on Flickr SoundNet, and test the Flickr trained model on the VGG-SS test set. 
As can be seen in Table~\ref{tab: exp_cross}, our approach outperforms the best previous method~\cite{chen2021localizing} when testing across datasets. 

\begin{table}[!tb]
	\renewcommand\tabcolsep{6.0pt}
	\centering
	\caption{Cross dataset generalization results of Flickr SoundNet and VGG-SS trained on various training sets, including VGG-Sound 10k, 144k, Full and Flickr 10k, 144k.}
	\label{tab: exp_cross}
	\scalebox{0.9}{
		\begin{tabular}{cclcc}
			\toprule
			Test set & Training set & Method & CIoU(\%) & AUC(\%)  \\ 	
			\midrule
             \multirow{6}{*}{Flickr SoundNet} & \multirow{2}{*}{VGG-Sound 10k} & LVS~\cite{chen2021localizing} & 61.80 & 53.60 \\
             & & EZ-VSL & \textbf{78.71} & \textbf{61.56} \\
             & \multirow{2}{*}{VGG-Sound 144k} & LVS~\cite{chen2021localizing} & 71.90 & 58.20 \\
             & & EZ-VSL & \textbf{84.34} & \textbf{63.77} \\
             & \multirow{2}{*}{VGG-Sound Full} & LVS~\cite{chen2021localizing} & 73.59 & 59.00 \\
             & & EZ-VSL & \textbf{83.94} & \textbf{63.60} \\ \hline
             \multirow{4}{*}{VGG-SS} & \multirow{2}{*}{Flickr 10k} & LVS~\cite{chen2021localizing} & 18.71 & 30.29 \\
             & & EZ-VSL & \textbf{35.54} & \textbf{38.18} \\
              & \multirow{2}{*}{Flickr 144k} & LVS~\cite{chen2021localizing} & 26.95 & 34.30\\
             & & EZ-VSL & \textbf{38.62} & \textbf{39.20}\\
             \bottomrule
			\end{tabular}}
\end{table}

\subsection{Experimental analysis}
We conducted extensive ablation studies to explore the benefits of the two main components of our approach: multiple instance contrastive learning (MICL) and object-guided localization (OGL).
We also conducted several parametric studies to assess the impact of hyper-parameters such as the size of shared audio-visual latent space, the audio-visual fusion strategy, or the balancing coefficient $\alpha$ used for OGL.
All experiments were trained on the VGG-Sounds full training set and evaluated on Flickr-SoundNet and VGG-Sound Source (VGG-SS) test sets.

\paragraph{\textbf{Disentangling the benefits of MICL and OGL.}}
We ablated the use of MICL and OGL to verify their effectiveness. Models evaluated without MICL only use the object guided localization maps extracted from the pre-trained ResNet-18, without any further training. Models evaluated without OGL only use the audio-visual localization (AVL) maps learned using MICL.
We further evaluate two strategies for OGL, namely, classification based OGL (CLS-OGL) described in~(\ref{eq:s_cls}) and activation based OGL (L1-OGL) described in (\ref{eq:s_l1}).

Results are shown in Table~\ref{tab: ab_obj_rce}.
First, comparing the performance of each component evaluated in isolation (first three rows of Table~\ref{tab: ab_obj_rce}) to those in Table~\ref{tab: exp_sota_vggss}, we highlight that both AVL and L1-OGL already surpass prior state-of-the-art (LVS~\cite{chen2021localizing}). Especially interesting is the observation that L1-OGL, which does not use any audio information, already achieves very strong performance by itself. This is likely due to two reasons. First, object regions are more likely to depict the sound sources. Second, the majority of test samples in both Flickr and VGG-SS only contain a single sounding object in the scene. This is more prevalent in Flickr but is still true for VGG-SS. As a result, only finding objects already provides strong localization results, outperforming all prior work.
Our framework can nevertheless improve over OGL, by combining it with audio-visual localization (AVL).
\begin{table}[!tb]
	\renewcommand\tabcolsep{6.0pt}
	\centering
	\caption{Ablation study on the impact of audio-visual localization (AVL) maps and two object-guided localization strategies (CLS and L1 prior) during inference. }
	\label{tab: ab_obj_rce}
	\scalebox{0.85}{
		\begin{tabular}{cccccccc}
			\toprule
			\multirow{2}{*}{AVL} & \multirow{ 2}{*}{L1-OGL} & \multirow{ 2}{*}{CLS-OGL} & \multicolumn{2}{c}{Flickr SoundNet} & \multicolumn{2}{c}{VGG-SS} \\
			& & & CIoU(\%) & AUC(\%) & CIoU(\%) & AUC(\%) \\ 	
			\midrule
			\checkmark & & &78.31 & 61.74 & 35.96 & 38.20 \\
			& \checkmark & &  78.31 & 61.17 & 36.77 & 38.69 \\
			& & \checkmark & 75.10 & 58.18 & 35.13 & 38.08 \\
			\checkmark & & \checkmark & 81.93 & 62.50 & 38.58 & 39.59\\
			 \checkmark & \checkmark & & \textbf{83.94} & \textbf{63.60} & \textbf{39.34} & \textbf{39.78} \\
			\bottomrule
			\end{tabular}}
\end{table}

\begin{figure*}[!tb]
	    \centering
	\includegraphics[width=\linewidth]{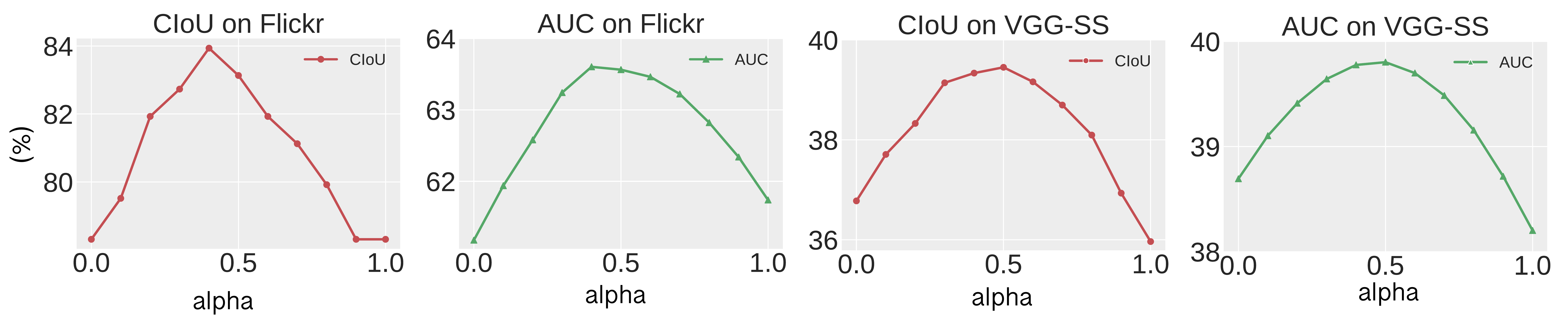}
  \caption{Effect of $\alpha$ on the object guided localization of our EZ-VSL for Flickr SoundNet and VGG-SS.}
	\label{fig: vis_alpha}
\end{figure*}

Among the two OGL strategies, L1-OGL was the most effective, and thus used as the default strategy for EZ-VSL. We also evaluated the localization performance for various values of the balancing coefficient $\alpha$ between AVL and L1-OGL localization maps.
The results in Fig.~\ref{fig: vis_alpha} show that both OGL and AVL components are important for accurate localization, as $\alpha=0$ or $\alpha=1$ yields the worse performance. The optimal value of $\alpha$ for Flickr was 0.4 and for VGG-SS was 0.5. $\alpha=0.4$ was used as the default for all experiments in this paper.

\begin{figure*}[!tb]
	    \centering
	\includegraphics[width=\linewidth]{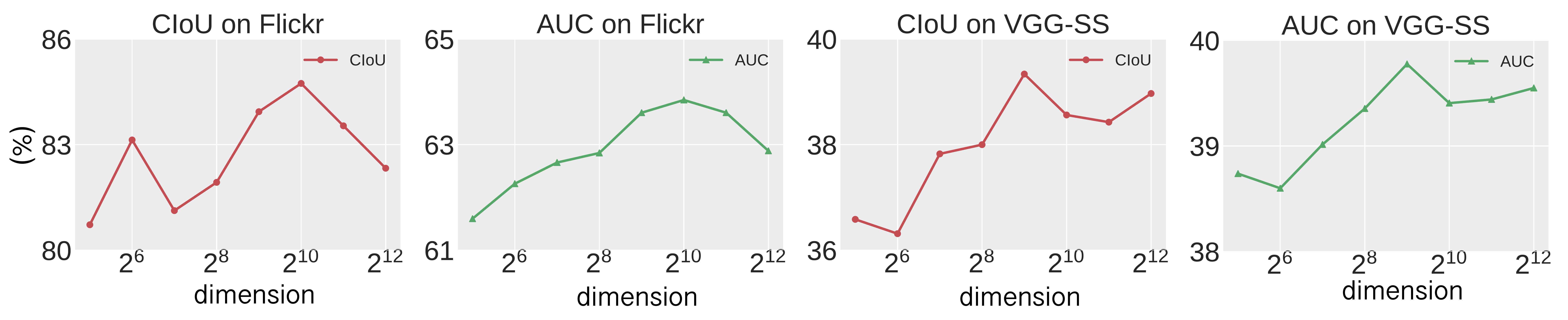}
    \caption{Effect of different output sizes on the final performance of our EZ-VSL model for Flickr SoundNet and VGG-SS testset. Optimal dimensionality is 1024 on Flickr SoundNet and 512 for VGG-SS.}
	\label{fig: vis_outputsize}
\end{figure*}

\paragraph{\textbf{Dimensionality of shared audio-visual latent space.}}
We show the impact of the latent space dimentionality in Fig.~\ref{fig: vis_outputsize}.
The models were trained on VGG-Sound with latent space of size 32, 64, 128, 256, 512, 1024, 2048, 4096, and tested on Flickr SoundNet and VGG-SS. It should be noted that both audio and visual encoders output 512-d representations. Fig.~\ref{fig: vis_outputsize} shows that significantly reducing or increasing the unimodal feature dimensionality can have a negative impact on performance. 

\begin{table}[!tb]
	\renewcommand\tabcolsep{6.0pt}
	\centering
	\caption{Impact of different audio-visual matching strategies during training on audio-visual localization performance. $V_{xy}$ denotes the visual embedding at location $(x, y)$, and $\mathbf{A}$ the global audio embedding. Only the audio-visual localization maps are evaluated in this experiment, without being merged with object-guided localization maps.}
	\label{tab: ab_pool}
	\scalebox{0.85}{
		\begin{tabular}{cccccc}
			\toprule
			\multirow{2}{*}{AV matching strategy} & \multicolumn{2}{c}{Flickr SoundNet} & \multicolumn{2}{c}{VGG-SS} \\
			& CIoU(\%) & AUC(\%) & CIoU(\%) & AUC(\%) \\ 
			\midrule
			$\mathtt{sim}(\mbox{MaxPool}_{xy}(V_{xy}), A)$
		    & 49.40	& 48.97 & 12.72 & 27.10\\
		    $\mbox{AvgPool}_{xy}(\mathtt{sim}(V_{xy}, A))$ & 33.33 & 37.56 & 6.03 & 19.44\\
			$\mbox{MaxPool}_{xy}(\mathtt{sim}(V_{xy}, A))$ & \textbf{78.31} & \textbf{61.74} & \textbf{35.96} & \textbf{38.20} \\
			\bottomrule
			\end{tabular}}
\end{table}

\paragraph{\textbf{Audio-visual matching strategy during training.}}
The proposed EZ-VSL method uses a max pooling strategy for measuring the similarity between the global audio feature $A$ and the bag of localized visual features $V=\{V_{xy}: \forall x, y\}$, i.e., using $\mathtt{sim}(\mbox{MaxPool}_{xy}(V_{xy}), A)$. 
We validate this strategy by comparing two alternatives. 
First, average pooling is a popular strategy for gathering responses across instances in a bag~\cite{Ilse2018AttentionbasedDM}. Therefore, we follow this approach and train a model that seeks to match the global audio feature to the visual features at \textit{all} locations, i.e., using $\mathtt{sim}(\mbox{AvgPool}_{xy}(V_{xy}), A)$. 
Second, prior work on audio-visual representation learning~\cite{Arandjelovic2017look,owens2016ambient,korbar2018cooperative,Morgado2021audio} learn by matching global features. We also tested this class of methods by training a model that pools the visual features before matching to the audio, i.e., using $\mathtt{sim}(\mbox{MaxPool}_{xy}(V_{xy}), A)$.

The localization performance of all three strategies  are reported in Table~\ref{tab: ab_pool}. Since only audio-visual localization maps are impacted by the different training strategies, we set $\alpha=1$ in this experiment to ignore object-guided localization maps.
As can be seen, the two alternative strategies failed to localize sounding objects accurately. 
On one hand, matching global features lacks the ability to learn localized representations. 
On the other hand, forcing the audio to match the image at all locations is also inherently problematic, since most regions do not contain a sounding object.
The proposed approach achieves significantly better localization performance. However, it assumes that there is at least one sounding object visible in the image. While this is generally true in both VGG-Sound and Flickr SoundNet training sets, further experiments on datasets with non-visible sound sources would be required to assess the robustness of EZ-VSL to this more challenging training scenario.

\subsection{Qualitative Results}
\begin{figure*}[!tb]
    \centering
    \includegraphics[width=\linewidth]{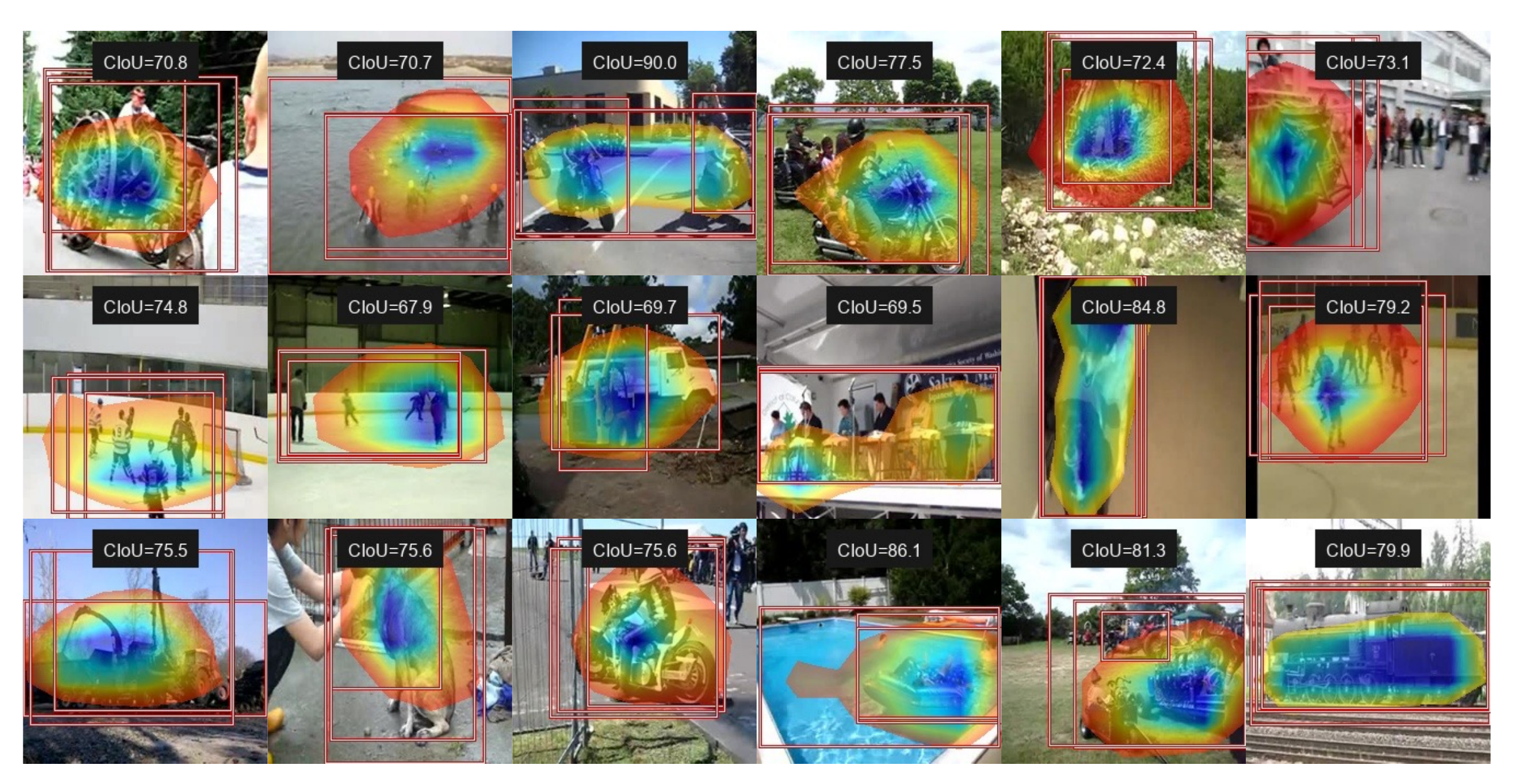}
    \caption{Predicted localization maps on Flickr SoundNet test images.}
	\label{fig: vis_good}
\end{figure*}

\begin{figure*}[!tb]
    \centering
    \includegraphics[width=\linewidth]{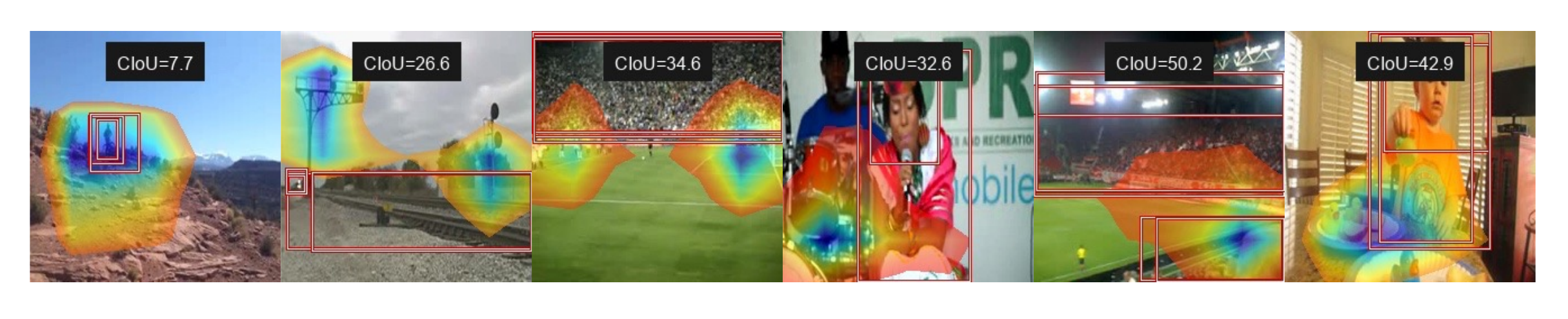}
    \caption{Failure cases of EZ-VSL. Typical cases which EZ-VSL still struggles to accurately localize sound sources include small objects, or when sounds are not produced by objects, such as the sound of crowds.}
	\label{fig: vis_bad}
\end{figure*}

To better understand the capabilities of the learned model, we show in Fig.~\ref{fig: vis_good} sound localization predictions of an EZ-VSL model trained on the VGG-SS 144k dataset.
As can be seen, the model is capable of accurately localizing a wide variety of sound sources, showing high overlap with the ground-truth bounding boxes. For example, in row 2, column 4, the model was able to identify that the sound sources are the musical instruments and not the people playing them, or that the sound source in row 3 column 2 is the dog (and not the man).
We also show failure cases in Fig.~\ref{fig: vis_bad}. We notice that the learned model often has trouble predicting tight localization maps for small objects, or localizing the sound of crowds, such as in stadiums.


\begin{figure*}[!tb]
	    \centering
	\includegraphics[width=\linewidth]{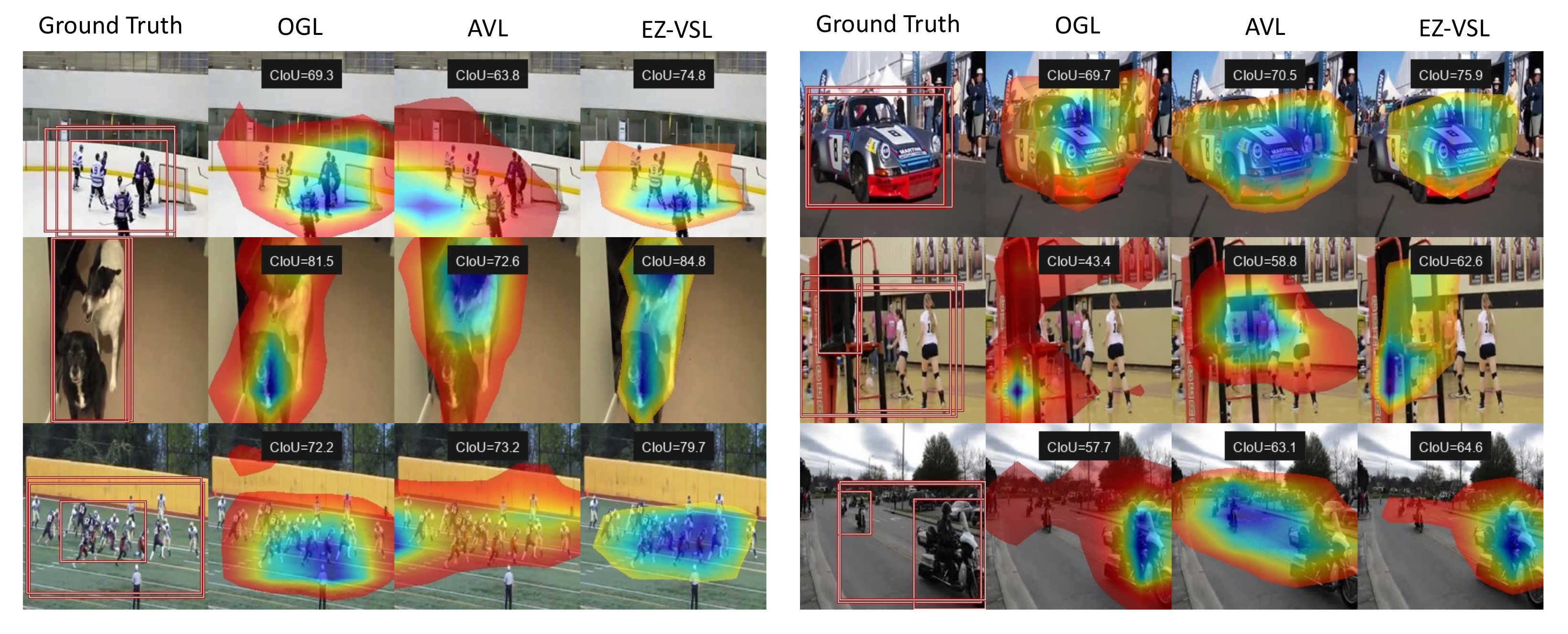}
  \caption{Sound source localization by OGL, AVL, and our EZ-VSL. Object-guided maps tend to cover all objects in the scene; audio-video similarity maps often cover the sounding object and some non-object regions; the final EZ-VSL map tends to better focus on the sounding object.}
	\label{fig: vis_alpha_map}
\end{figure*}

Finally, we compare the final localization map with the object-guided map and the audio-visual similarity map in Fig.~\ref{fig: vis_alpha_map}.
These results demonstrate the effectiveness of combining object-guided and audio-visual localization in visual sound localization.

\section{Conclusion}
\label{sec:conclusion}

In this work, we present the EZ-VSL, a simple yet effective approach for visual sounds source localization, with no need to explicitly compute the negative regions.
Specifically, a simple cross-entropy loss is applied to learn the relative correspondence between the visual and audio instances.
Furthermore, we propose a novel object-guided localization scheme to mix the audio-visual joint map and the object map from a lightweight pre-trained visual model for boosting the performance of orientating sound sources in an image.
Compared to previous contrastive and non-contrastive baselines, our framework successfully achieves state-of-the-art performance on two popular benchmarks, Flickr SoundNet and VGG-Sound Source.
Comprehensive ablation studies are conducted to show the effectiveness of each component in our simple method.
We also demonstrate the significant advantage of our approach on the open set visual sounds source localization and cross dataset generalization.

%
%
\bibliographystyle{splncs04}
\bibliography{references}
\end{document}